\documentclass[preprint,12pt]{elsarticle}

\usepackage{amssymb}
\usepackage{url}
\usepackage{subfigure}
\usepackage{amsmath}
\usepackage{threeparttable}
\usepackage{algorithm}
\usepackage{algorithmic}
\usepackage{epstopdf}
\usepackage{booktabs}
\usepackage{color}
\usepackage{multirow}

\newtheorem{prop}{Proposition}
\biboptions{numbers,sort&compress}

\usepackage{lineno}

\begin{document}

\begin{frontmatter}

\title{Robust Bhattacharyya bound linear discriminant analysis through adaptive algorithm}
\tnotetext[label1]{This work is supported by the National Natural Science Foundation of China (No.61703370, No.61866010, No.11871183, No.11501310 and No.61603338), the Natural Science Foundation of Zhejiang Province (No.LQ17F030003 and No.LY18G010018), and the Hainan Provincial Natural Science Foundation of China (No.118QN181).}



\author[1]{Chun-Na Li}
\address[1]{Zhijiang College, Zhejiang University of Technology, Hangzhou,
310024, P.R.China}

\author[2]{Yuan-Hai Shao}
\address[2]{School of Economics and Management, Hainan University, Haikou, 570228, P.R.China.}

\author[3]{Zhen Wang}
\address[3]{School of Mathematical Sciences, Inner Monggolia University, Hohhot, 010021, P.R.China}

\author[4]{Nai-Yang Deng}
\address[4]{College of Science, China Agricultural University, Beijing, 100083, P.R.China}

\begin{abstract}
In this paper, we propose a novel linear discriminant analysis criterion via the Bhattacharyya error bound estimation based on a novel L1-norm (L1BLDA) and L2-norm (L2BLDA). Both L1BLDA and L2BLDA maximize the between-class scatters which are measured by the weighted pairwise distances of class means and meanwhile minimize the within-class scatters under the L1-norm and L2-norm, respectively.
The proposed models can avoid the small sample size (SSS) problem and have no rank limit that may encounter in LDA. It is worth mentioning that, the employment of L1-norm gives a robust performance of L1BLDA, and L1BLDA is solved through an effective non-greedy alternating direction method of multipliers (ADMM), where all the projection vectors can be obtained once for all. In addition, the weighting constants of L1BLDA and L2BLDA between the between-class and within-class terms are determined by the involved data set, which makes our L1BLDA and L2BLDA adaptive.
The experimental results on both benchmark data sets as well as the handwritten digit databases demonstrate the effectiveness of the proposed methods.
\end{abstract}
\begin{keyword}
dimensionality reduction; linear discriminant analysis; robust linear discriminant analysis; Bhattacharyya error bound; alternating direction method of multipliers
\end{keyword}

\end{frontmatter}


\section{Introduction}
Linear discriminant analysis (LDA) \cite{Fisher36,Fukunaga90} is a well-known supervised dimensionality reduction method, and has been extensively studied since it was proposed.
LDA tries to find an optimal linear transformation by maximizing the quadratic distance between the class means simultaneously minimizing the within-class distance in the projected space. Due to its simplicity and effectiveness, LDA is widely applied in many applications, including image recognition \cite{Belhumeur97,SunXie16,LuoHou18,LiShaoChen18}, gene expression \cite{GuoHastie06}, biological populations \cite{JombartDevillard10}, image retrieval \cite{SwetsWeng96}, etc.

Despite the popularity of LDA, there exist some drawbacks that restrict its applications.
As we know, LDA is solved through a generalized eigenvalue problem $\textbf{S}_b\textbf{w}=\lambda \textbf{S}_w\textbf{w}$, where $\textbf{S}_b$ and $\textbf{S}_w$ are the classical between-class scatter and the within-class scatter, respectively. When dealing with the SSS problem, $\textbf{S}_w$ is not of full rank and LDA will encounter the singularity.
Moreover, since LDA is constructed based on the L2-norm, it is sensitive to the presence of outliers. These make LDA non-robust. In addition, since the rank of $\textbf{S}_b$ is most $c-1$, where $c$ is the class number, LDA can find at most $c-1$ meaningful features, which is also a limitation.

For solving the above non-robustness issues, many endeavors were made from different aspects, including using the null space information \cite{ChenLiao00,YuYang01}, the subspace learning technique \cite{Belhumeur97,SwetsWeng96,LaiMo18}, the regularization technique \cite{Friedman89,GuoHastie06},
incorporating a model of data uncertainty in the classification and optimizing for the worst-case \cite{KimMagnaniBoyd06}, utilizing the pseudo inverse of $\textbf{S}_w$ \cite{TianBarbero86}, and using the robust mean and scatter variance estimators \cite{CrouxDehon01,HubertDriessen04}.
For the rank limit issue, incorporating the local information \cite{Sugiyama07} and the recursive technique \cite{XiangFan06,YeZhaoetc12,ChenYangetal13,LiZheng17} were usually considered. Recently, the employment of the L1-norm rather than the L2-norm in LDA was studied to cope with the non-robustness and rank limit problems.
Li et al. \cite{LiHuaWang10} considered a rotational invariant L1-norm (R1-norm) based LDA, while the L1-norm based LDA-L1 \cite{ZhongZhang13,WangLu14,Liu17}, ILDA-L1 \cite{ChenYangetal14}, L1-LDA \cite{Zheng14} and L1-ELDA \cite{YeYangLiu18} were also put forward, where their scatter matrices are measured by the R1-norm and L1-norm, respectively. The matrix based LDA-L1 was further raised and studied \cite{LiShao15,ChenL12DLDA15,TrL12DLDA17}.
The extension to the Lp-norm ($p>0$) \cite{KwakLpLDA,An14} scatter covariances was also used in LDA. However, as pointed in \cite{YeYangLiu18}, some of the above methods were still not robust enough.



As we know, minimizing the Bhattacharyya error \cite{Bhattacharyya43} bound is a reasonable way to establish classification \cite{Devijveretal82,Fukunaga90}. In this paper, based on the Bhattacharyya error bound, a novel robust L1-norm linear discriminant analysis (L1BLDA) and its corresponding L2-norm criterion (L2BLDA) are proposed. Both of them can avoid the singularity and the rank limit issues, 
and the employment of the L1-norm makes our L1BLDA more robust.



In summary, the proposed L1BLDA and L2BLDA have the following several characteristics:
\begin{figure*}[htpb]
\begin{center}{
\subfigure[Original data and the projection directions]{
\resizebox*{6.5cm}{!}
{\includegraphics{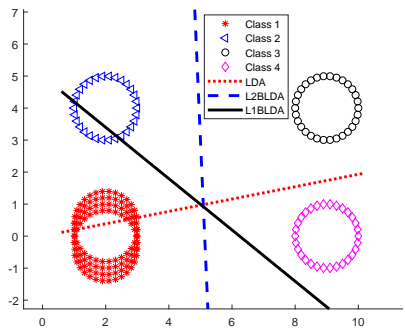}}}\hspace{5pt}
\subfigure[Projected data]{
\resizebox*{6.5cm}{!}
{\includegraphics{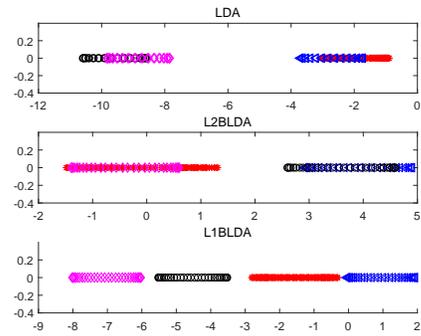}}}\hspace{5pt}
\subfigure[Data with outliers and the projection directions]{
\resizebox*{6.5cm}{!}
{\includegraphics{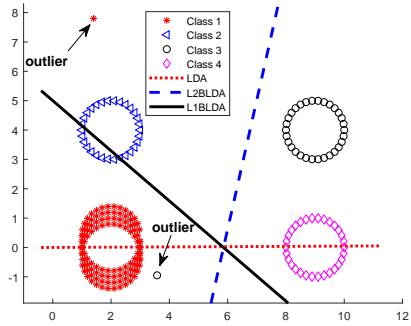}}}\hspace{5pt}
\subfigure[Projected data with outliers]{
\resizebox*{6.5cm}{!}
{\includegraphics{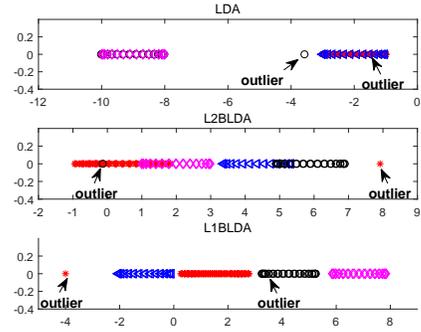}}}\hspace{5pt}
\caption{Artificial data set and its projections obtained by LDA, L2BLDA, and L1BLDA.}
\label{arti3}}
\end{center}
\end{figure*}

$\bullet$ Both L1BLDA and L2BLDA are derived by minimizing the Bhattacharyya error bound, which ensures the rationality of the proposed methods. Specifically, we prove that the upper bound of the Bhattacharyya error can be expressed linearly by the between-class scatter and the within-class scatter, so that minimizing this upper bound leads to our optimization problem of the form $\min~-\textbf{S}_{bB}+ D\cdot \textbf{S}_{wB}$, where $\textbf{S}_{bB}$ is the between-class scatter, $\textbf{S}_{wB}$ is the within-class scatter, and $D$ weights $\textbf{S}_{bB}$ and $\textbf{S}_{wB}$. In particular, it should be pointed out that the weight $D$ is calculated based the input data, so that our models adapt to different data sets.

$\bullet$ For the between-class scatter and within-class scatter, L1BLDA uses the L1-norm (LASSO) loss $f_1(a)=|a|$, while both the scatters of L2BLDA and LDA are described by the L2-norm (square) loss $f_2(a)=|a|^2$.
It is obvious that the difference between $f_2(a)$ and $f_1(a)$ becomes larger as $|a|$ getting larger, so we expect that L1BLDA is more robust than L2BLDA and LDA when the data set contains outliers.



To testify the robustness of L1BLDA, we here perform an experiment on a simple data set with four classes. The first class contains 120 data samples, while each of the other three classes contains 30 data samples. We apply the following three algorithms LDA, our L1BLDA and L2BLDA on the data set and obtain the one dimensional projected data, as shown in Fig. \ref{arti3}. Then two additional outliers are added on the above data for testing.
It is obvious that for our L1BLDA, the outliers have little influence to its projection direction, and the projected samples are separated well. On the contrary, LDA and LB2DLDA are greatly affected by outliers.

$\bullet$ Two nongreedy adaptive algorithms are proposed for the optimization problems in solving L1BLDA and L2BLDA, respectively: i) for L1BLDA, it is solved by an effective ADMM algorithm, which is characterized by a one-time projection matrix without the need to recursively solve a single projection vector. Compared with traditional recursive algorithm for L1-norm based LDA, our ADMM approach could maintain the orthogonality and the normalization of the projection directions; ii) for L2BLDA, it is solved through a standard eigenvalue decomposition problem that does not involve the inversion operation, rather than a generalized eigenvalue decomposition problem in LDA.


$\bullet$ Our L1BLDA and L2BLDA can avoid the singularity caused by the SSS problem. Moreover, L1BLDA does not have the rank limit issue.


%


The notation of the paper is given as follows. All vectors are column ones. Given the training set $T=\{\textbf{x}_1,\,\textbf{x}_2,\ldots,\,\textbf{x}_N\}$ with the associated class labels $y_1,\,y_2,\ldots,\,y_N$ belonging to $\{1,\,2,\,\ldots,c\}$, where $\textbf{x}_l\in\mathbb{R}^n$ for $l=1,\,2,\,\ldots,N$.
Denote $\textbf{X}=(\textbf{x}_1,\,\textbf{x}_2,\ldots,\,\textbf{x}_N)\in\mathbb{R}^{n\times N}$ as the data matrix.
Assume that the $i$-th class contains $N_{i}$ samples. Then $\sum\limits_{i=1}^{c}N_i=N$. Let $\overline{\textbf{x}}=\frac{1}{N}\sum\limits_{l=1}^{N}\textbf{x}_l$ be the mean of all samples and ${\overline{\textbf{x}}}_i=\frac{1}{N_i}\sum\limits_{j=1}^{N_i}\textbf{x}_{ij}$ be the mean of samples in the $i$-th class.

The paper is organized as follows. Section \ref{L2LDA} briefly reviews LDA. Section \ref{secL2BLDA} and Section \ref{secL1BLDA} elaborate on our L2BLDA and L1BLDA, respectively. Section \ref{secExp} makes comparisons of the proposed methods with their related methods. At last, concluding remarks are given in Section \ref{secCon}.

\section{Linear discriminant analysis}\label{L2LDA}

The main idea of LDA is to find an optimal projection transformation matrix $\textbf{W}$ such that the ratio of between-class scatter to within-class scatter is maximized in the projected space of $\textbf{W}\in\mathbb{R}^{n\times d}$, $d\leq n$. Specifically, LDA solves the following optimal problem
\begin{equation}\label{LDA}
\begin{split}
\underset{\textbf{W}}{\max}&~~\frac{\hbox{tr}(\textbf{W}^T\textbf{S}_b\textbf{W})}{\hbox{tr}(\textbf{W}^T\textbf{S}_w\textbf{W})},\\
\end{split}
\end{equation}
where the between-class scatter matrix $\textbf{S}_b$ and the within-class scatter matrix $\textbf{S}_w$ are defined by
\begin{equation}\label{SB}
\textbf{S}_b=\frac{1}{N}\sum\limits_{i=1}^{c}N_i({\overline{\textbf{x}}}_i-{\overline{\textbf{x}}})({\overline{\textbf{x}}}_i-{\overline{\textbf{x}}})^T
\end{equation}
and
\begin{equation}\label{SW} \textbf{S}_w=\frac{1}{N}\sum\limits_{i=1}^{c}\sum\limits_{j=1}^{N_i}(\textbf{x}_{ij}-{\overline{\textbf{x}}_i})(\textbf{x}_{ij}-{\overline{\textbf{x}}_i})^T.
\end{equation}
The optimization problem \eqref{LDA} is equivalent to the generalized problem $\textbf{S}_b\textbf{w}=\lambda \textbf{S}_w\textbf{w}$ where $\lambda\not=0$, with its solution $\textbf{W}=(\textbf{w}_1,\ldots,\textbf{w}_d)$ given by the first $d$ largest eigenvalues of $(\textbf{S}_w)^{-1}\textbf{S}_b$ in case $\textbf{S}_w$ is nonsingular. Since the rank of $\textbf{S}_b$ is at most $c-1$, the number of extracted features is less or equal than $c-1$.
\section{L2-norm linear discriminant analysis criterion via the Bhattacharyya error bound estimation}\label{secL2BLDA}

The error probability minimization is a natural way to obtain dimensionality reduction for classification, who involves the maximization of probabilistic distance measures and probabilistic dependence measures between different classes \cite{TouHeydorn67,Fukunaga90,Devijveretal82,Belhumeur97,DuinLoog04,TorreKanade05,AbouTorre10}.
Since the Bayes classifier is the best classifier which minimizes the probability of classification error, minimizing its error rate (called the Bayes error or probability of misclassification \cite{Fukunaga90}) is expected to lead to good classification model. The Bayes error is defined as
\begin{equation}\label{bayesoo}
\epsilon = 1- \int\underset{i\in\{1,2,\ldots,c\}}{max}\{P_ip_i(\textbf{x})\}d\textbf{x},
\end{equation}
where $\textbf{x}\in \mathbb{R}^n$ is a sample vector, $P_i$ and $p_i(\textbf{x})$ are the prior probability and the probability density function of the $i$th class for the data set $T$, respectively.
The computation of the Bayes error is a very difficult task except in some special cases, and an alternative way of minimizing the Bayes error is to minimize its upper bound \cite{Fukunaga90,Saon02}.
In particular, the Bhattacharyya error \cite{Bhattacharyya43} is a widely used upper bound that provides a close bound to the Bayes error.
In the following, we will derive a novel L2-norm linear discriminant analysis criterion via the Bhattacharyya error bound estimation, named L2BLDA, and give its solving algorithm.
The Bhattacharyya error bound is given by
\begin{equation}\label{Berror}
\epsilon_B=\sum\limits_{i<j}^c\sqrt{P_iP_{j}}\int\sqrt{p_i(\textbf{x})p_{j}(\textbf{x})}d\textbf{x}.\\
\end{equation}
We now derive an upper bound of $\epsilon_B$ under some assumptions.
\begin{prop}\label{propL2}
Assume $P_i$ and $p_i(\textbf{x})$ are the prior probability and the probability density function of the $i$th class for the training data set $T$, respectively, and the data samples in each class are independent identically normally distributed. Let $p_1(\textbf{x}), p_2(\textbf{x}),\ldots, p_c(\textbf{x})$ be the Gaussian functions given by $p_i(\textbf{x})=\mathcal{N}(\textbf{x}|{\overline{\textbf{x}}}_i, \boldsymbol{\Sigma}_i)$, where ${\overline{\textbf{x}}}_i$ and $\boldsymbol{\Sigma}_i$ are the class mean and the class covariance matrix, respectively. We further suppose $\boldsymbol{\Sigma}_i=\boldsymbol{\Sigma}$, $i=1,2,\ldots,c$, and ${\overline{\textbf{x}}}_i$ and $\boldsymbol{\Sigma}_i$ can be estimated accurately from the training data set $T$. Then for arbitrary projection vector $\textbf{w}\in\mathbb{R}^n$,
the Bhattacharyya error bound $\epsilon_B$ defined by \eqref{Berror} on the data set $\widetilde{T}=\{\widetilde{x}_i|\widetilde{x}_i=\textbf{w}^T\textbf{x}_i\}$ satisfies the following:
\begin{equation}\label{BhattacharyyaL2}
\begin{split}
\epsilon_B
\leq&-\frac{N}{8}\sum\limits_{i<j}^c\sqrt{P_iP_j}{||\textbf{w}^T({\overline{\textbf{x}}_{i}-\overline{\textbf{x}}_{j}})||_2^2}+\frac{N}{8}\Delta{||\textbf{w}^T\textbf{X}-\textbf{w}^T\overline{\textbf{x}}_{\textbf{I}}||_2^2}\\
&+\sum\limits_{i<j}^c\sqrt{P_iP_j},\\
\end{split}
\end{equation}
where $\Delta=\frac{1}{4}\sum\limits_{i<j}^c\sqrt{P_iP_j}||{\overline{\textbf{x}}_{i}-\overline{\textbf{x}}_{j}}||_2^2$.
\end{prop}

\noindent \textbf{Proof:} We first note that $p_i(\widetilde{x})=\mathcal{N}(\widetilde{x}|\widetilde{{\overline x}}_i, \sigma_i)$, where $\widetilde{{\overline x}}_i=\textbf{w}^T\widetilde{\textbf{x}}_i$ is the $i$-class mean and $\sigma_i$ is the $i$-class standard variance in the 1-dimensional projected space.

Then we have \cite{Fukunaga90}
\begin{equation}\label{Bbound}
\begin{split}
\int\sqrt{p_i(\widetilde{x})p_{j}(\widetilde{x})}= e^{-\frac{(\widetilde{{\bar x}_{i}}-\widetilde{{\bar x}_{j}})^2}{8\sigma^2}}.
\end{split}
\end{equation}

Since
\begin{equation}\label{variance}
\begin{split}
\sigma^2=\frac{1}{N}\sum_{i=1}^{c}\sum_{j=1}^{N_i}(\textbf{w}^T(\textbf{x}_j^i-{\overline{\textbf{x}}}_i))_2^2=&{\frac{1}{N}\sum\limits_{l=1}^{N}(\textbf{w}^T\textbf{x}_l-\textbf{w}^T\overline{\textbf{x}}_{t_l})^2}\\
=&\frac{1}{N}{||\textbf{w}^TX-\textbf{w}^T\overline{\textbf{x}}_{\textbf{I}}||_2^2},
\end{split}
\end{equation}
we have
\begin{equation}\label{Bhattacharyya1}
\begin{split}
\epsilon_B
&\leq\sum\limits_{i<j}^c \sqrt{P_iP_j} e^{-\frac{1}{8}\frac{(\widetilde{{\bar x}_{i}}-\widetilde{{\bar x}_{j}})^2}{\sigma^2}}\\
&=\sum\limits_{i<j}^c \sqrt{P_iP_j} e^{-(\frac{\widetilde{{\bar x}_{i}}-\widetilde{{\bar x}_{j}}}{2\sqrt{2}\sigma})^2}\\
&\leq\sum\limits_{i<j}^c \sqrt{P_iP_j} (1-\big(\frac{\widetilde{{\bar x}_{i}}-\widetilde{{\bar x}_{j}}}{2\sqrt{2}\sigma}\big)^2)\\
&=\sum\limits_{i<j}^c \sqrt{P_iP_j} -\frac{N}{8}\sum\limits_{i<j}^c\sqrt{P_iP_j}\cdot \frac{||\textbf{w}^T({\overline{\textbf{x}}_{i}-\overline{\textbf{x}}_{j}})||_2^2}{||\textbf{w}^TX-\textbf{w}^T\overline{\textbf{x}}_{\textbf{I}}||_2^2}\\
&\leq\sum\limits_{i<j}^c \sqrt{P_iP_j} -\frac{N}{8}\sum\limits_{i<j}^c\sqrt{P_iP_j}\cdot ||\textbf{w}^T({\overline{\textbf{x}}_{i}-\overline{\textbf{x}}_{j}})||_2^2\\
&~~~+\frac{N}{8}\sum\limits_{i<j}^c\sqrt{P_iP_j}\cdot \Delta_{ij}'||\textbf{w}^TX-\textbf{w}^T\overline{\textbf{x}}_{\textbf{I}}||_2^2,
\end{split}
\end{equation}
where $\Delta_{ij}'\geq \frac{1}{4}||{\overline{\textbf{x}}_{i}-\overline{\textbf{x}}_{j}}||_2^2$, $\overline{\textbf{x}}_{\textbf{I}}=(\overline{\textbf{x}}_{t_1},\ldots,\overline{\textbf{x}}_{t_l},\ldots,\overline{\textbf{x}}_{t_N})\in \mathbb{R}^{n\times N}$, and $\overline{\textbf{x}}_{t_l}$ is the center of the class that the $l$-th sample $\textbf{x}_l$ belongs to, $l=1,2,\ldots,N$.
The second inequality of \eqref{Bhattacharyya1} holds by the fact that $e^{-x^2}\leq1-x^2$ for $x\geq0$.
For the last inequality, since $||\textbf{w}^T({\overline{\textbf{x}}_{i}-\overline{\textbf{x}}_{j}})||_2^2\leq||\textbf{w}||_2^2\cdot||{\overline{\textbf{x}}_{i}-\overline{\textbf{x}}_{j}}||_2^2 = ||{\overline{\textbf{x}}_{i}-\overline{\textbf{x}}_{j}}||_2^2$ and $\frac{1}{||\textbf{w}^T\textbf{X}-\textbf{w}^T\overline{\textbf{x}}_{\textbf{I}}||_2^2}\big(1-\frac{1}{||\textbf{w}^T\textbf{X}-\textbf{w}^T\overline{\textbf{x}}_{\textbf{I}}||_2^2}\big)\leq \frac{1}{4}$, we have
\begin{equation}\label{ueq1}
\begin{split}
&||\textbf{w}^T({\overline{\textbf{x}}_{i}-\overline{\textbf{x}}_{j}})||_2^2\cdot\frac{1}{||\textbf{w}^T\textbf{X}-\textbf{w}^T\overline{\textbf{x}}_{\textbf{I}}||_2^2}(1-\frac{1}{||\textbf{w}^T\textbf{X}-\textbf{w}^T\overline{\textbf{x}}_{\textbf{I}}||_2^2})\\
\leq &\frac{1}{4}||{\overline{\textbf{x}}_{i}-\overline{\textbf{x}}_{j}}||_2^2,
\end{split}
\end{equation}
which implies
\begin{equation}\label{ueq2}
\begin{split}
&\frac{||\textbf{w}^T({\overline{\textbf{x}}_{i}-\overline{\textbf{x}}_{j}})||_2^2}{||\textbf{w}^T\textbf{X}-\textbf{w}^T\overline{\textbf{x}}_{\textbf{I}}||_2^2}\\
\geq&||\textbf{w}^T({\overline{\textbf{x}}_{i}-\overline{\textbf{x}}_{j}})||_2^2-\frac{1}{4}||{\overline{\textbf{x}}_{i}-\overline{\textbf{x}}_{j}}||_2^2\cdot ||\textbf{w}^T\textbf{X}-\textbf{w}^T\overline{\textbf{x}}_{\textbf{I}}||_2^2\\
\geq&||\textbf{w}^T({\overline{\textbf{x}}_{i}-\overline{\textbf{x}}_{j}})||_2^2-\Delta_{ij}'\cdot ||\textbf{w}^T\textbf{X}-\textbf{w}^T\overline{\textbf{x}}_{\textbf{I}}||_2^2,
\end{split}
\end{equation}
and hence \eqref{Bhattacharyya1}.

By taking $\Delta=\sum\limits_{i<j}^c\sqrt{P_iP_j}\Delta_{ij}'= \frac{1}{4}\sum\limits_{i<j}^c\sqrt{P_iP_j}||{\overline{\textbf{x}}_{i}-\overline{\textbf{x}}_{j}}||_2^2$, we then obtain \eqref{BhattacharyyaL2}. \hfill$\square$

We are now ready to derive our optimazation problem from Proposition \ref{propL2}.
As stated before, we try to minimize the Bhattacharyya error bound. Therefore, by minimizing the right side of \eqref{BhattacharyyaL2} and neglecting the third constant term $\sum\limits_{i<j}^c \sqrt{P_iP_j}$ and the constant coefficient $\frac{N}{8}$, we get the formulation of our L2-norm based Bhattacharyya bound linear discriminant analysis (L2BLDA) as
\begin{equation}
\begin{split}\label{L2BLDAw}
\underset{\textbf{w}}{\min}~~&-\frac{1}{N}\sum_{i<j}\sqrt{N_iN_j}|\textbf{w}^T(\overline{\textbf{x}}_{i}-\overline{\textbf{x}}_{j})|^2+\Delta\sum_{i=1}^{c}\sum_{j=1}^{N_i}|\textbf{w}^T(\textbf{x}_j^i-\overline{\textbf{x}}_i)|^2\\
\hbox{s.t.\ }& \textbf{w}^T\textbf{w}=1,
\end{split}
\end{equation}
where $\Delta=\frac{1}{4}\sum\limits_{i<j}^c\sqrt{P_iP_j}||{\overline{\textbf{x}}_{i}-\overline{\textbf{x}}_{j}}||_2^2$.
The above optimization problem gives us one projective discriminant direction. In general, to project the data into higher dimensional space, our L2BLDA formulates as the following:
\begin{equation}
\begin{split}\label{L2BLDA}
\underset{\textbf{W}}{\min}~~&-\frac{1}{N}\sum_{i<j}\sqrt{N_iN_j}||\textbf{W}^T(\overline{\textbf{x}}_{i}-\overline{\textbf{x}}_{j})||_2^2+\Delta\sum_{i=1}^{c}\sum_{j=1}^{N_i}||\textbf{W}^T(\textbf{x}_j^i-\overline{\textbf{x}}_i)||_2^2\\
\hbox{s.t.\ }& \textbf{W}^T\textbf{W}=\textbf{\textbf{I}},
\end{split}
\end{equation}
where $\textbf{W}\in\mathbb{R}^{n\times d}$, $d\leq n$.
The geometric meaning of L2BLDA is clear.
By minimizing the first term in \eqref{L2BLDA}, the sum of the distances in the projected space between the centroid of $i$-th class and the centroid not in the $i$-th class is guaranteed to be as large as possible.
Minimizing the second term in \eqref{L2BLDA} makes sure any sample be close to its own class centroid in the low dimensional space. The coefficients $\frac{1}{N}\sqrt{N_iN_j}$ in the first term weight distance pairs between the $i$-th and the $j$-th class, while the constant $\Delta$ in front of the second term plays the balancing role while also makes sure minimum error bound be guaranteed.
The constraint $\textbf{W}^T\textbf{W}=\textbf{I}$ forces the obtained discriminant directions orthogonormal to each other, which ensures minimum redundancy in the projected space.

L2BLDA can be solved through a simple standard eigenvalue decomposition problem. In fact,
problem \eqref{L2BLDA} can be rewritten as
\begin{equation}
\begin{split}\label{L2BLDA2}
\underset{\textbf{W}}{\min}&~~tr(\textbf{W}^T\textbf{S}\textbf{W})\\
\hbox{s.t.\ }& \textbf{W}^T\textbf{W}=\textbf{I},
\end{split}
\end{equation}
where
\begin{equation}
\begin{split}
\textbf{S}=&-\frac{1}{N}\sum_{i<j}\sqrt{N_iN_j}(\overline{\textbf{x}}_{i}-\overline{\textbf{x}}_{j})(\overline{\textbf{x}}_{i}-\overline{\textbf{x}}_{j})^T+\Delta\sum_{i=1}^{c}\sum_{j=1}^{N_i}(\textbf{x}_j^i-\overline{\textbf{x}}_i)(\textbf{x}_j^i-\overline{\textbf{x}}_i)^T.
\end{split}
\end{equation}
Then $\textbf{W}=(\textbf{w}_1,\textbf{w}_2,\ldots,\textbf{w}_d)$ is given by the $d$ orthogonormal eigenvectors that correspond to the first $d$ smallest eigenvectors of $\textbf{S}$.

\section{L1-norm linear discriminant analysis criterion via the Bhattacharyya error bound estimation}\label{secL1BLDA}

\subsection{L1BLDA Bhattacharyya error bound derivation}\label{BhattL1B2DLDA}

In this section, we derive a different upper bound of the Bhattacharyya error under the L1-norm measure, aiming to construct a robust L1-norm based Bhattacharyya bound linear discriminant analysis (L1BLDA). Similar to L2BLDA, we first give the following proposition.
\begin{prop}
Assume $P_i$ and $p_i(\textbf{x})$ are the prior probability and the probability density function of the $i$th class for the training data set $T$, respectively, and the data samples in each class are independent identically normally distributed. Let $p_1(\textbf{x}), p_2(\textbf{x}),\ldots, p_c(\textbf{x})$ be the Gaussian functions given by $p_i(\textbf{x})=\mathcal{N}(\textbf{x}|{\overline{\textbf{x}}}_i, \boldsymbol{\Sigma}_i)$, where ${\overline{\textbf{x}}}_i$ and $\boldsymbol{\Sigma}_i$ are the class mean and the class covariance matrix, respectively. We further suppose $\boldsymbol{\Sigma}_i=\boldsymbol{\Sigma}$, $i=1,2,\ldots,c$, and ${\overline{\textbf{x}}}_i$ and $\boldsymbol{\Sigma}_i$ can be estimated accurately from the training data set $T$. Then for arbitrary projection vector $\textbf{w}\in\mathbb{R}^n$, there exist some constants $B$ and $C$ such that
the Bhattacharyya error bound $\epsilon_B$ defined by \eqref{Berror} on the data set $\widetilde{T}=\{\widetilde{x}_i|\widetilde{x}_i=\textbf{w}^T\textbf{x}_i\}$ satisfies the following:
\begin{equation}\label{BhattacharyyaL1}
\begin{split}
\epsilon_B
\leq&-B\sum\limits_{i<j}^c\sqrt{P_iP_j}{||\textbf{w}^T({\overline{\textbf{x}}_{i}-\overline{\textbf{x}}_{j}})||_1}+B\Omega{||\textbf{w}^T\textbf{X}-\textbf{w}^T\overline{\textbf{x}}_{\textbf{I}}||_1}\\
&+C\sum\limits_{i<j}^c\sqrt{P_iP_j},\\
\end{split}
\end{equation}
where $\Omega=\frac{\sqrt{n}}{4}\sum\limits_{i<j}^c\sqrt{P_iP_j}||{\overline{\textbf{x}}_{i}-\overline{\textbf{x}}_{j}}||_1$.
\end{prop}

\noindent{\textbf{Proof:}}
From \eqref{Bbound}, we have
\begin{equation}\label{Bhattacharyya0L1}
\begin{split}
\epsilon_B
&\leq\sum\limits_{i<j}^c \sqrt{P_iP_j} e^{-\frac{1}{8}\frac{(\widetilde{\overline{\textbf{x}}_{i}}-\widetilde{\overline{\textbf{x}}_{j}})^2}{\sigma^2}}\\
&\leq\sum\limits_{i<j}^c \sqrt{P_iP_j} e^{-\frac{1}{8}\frac{(\textbf{w}^T{\overline{\textbf{x}}_{i}}-\textbf{w}^T{\overline{\textbf{x}}_{j}})^2}{\frac{1}{N}\sum\limits_{i=1}^{N}(\textbf{w}^T\textbf{x}_k-\textbf{w}^T\overline{\textbf{x}}_{l_k})^2}}\\
&=\sum\limits_{i<j}^c  \sqrt{P_iP_j} e^{-\frac{N}{8}\big(\frac{||\textbf{w}^T({\overline{\textbf{x}}_{i}-\overline{\textbf{x}}_{j}})||_2}{||\textbf{w}^T\textbf{X}-\textbf{w}^T\overline{\textbf{x}}_{\textbf{I}}||_2}\big)^2}\\
&\leq\sum\limits_{i<j}^c  \sqrt{P_iP_j} e^{-\frac{N}{8}\big(\frac{||\textbf{w}^T({\overline{\textbf{x}}_{i}-\overline{\textbf{x}}_{j}})||_1}{||\textbf{w}^T\textbf{X}-\textbf{w}^T\overline{\textbf{x}}_{\textbf{I}}||_1}\big)^2}.\\
\end{split}
\end{equation}
Here
the last inequality of \eqref{Bhattacharyya0L1} follows from the fact that
\begin{equation}
\frac{||\textbf{w}^T({\overline{\textbf{x}}_{i}-\overline{\textbf{x}}_{j}})||_2}{||\textbf{w}^T\textbf{X}-\textbf{w}^T\overline{\textbf{x}}_{\textbf{I}}||_2}=\frac{||\textbf{w}^T({\overline{\textbf{x}}_{i}-\overline{\textbf{x}}_{j}})||_1}{||\textbf{w}^T\textbf{X}-\textbf{w}^T\overline{\textbf{x}}_{\textbf{I}}||_2}\geq \frac{||\textbf{w}^T({\overline{\textbf{x}}_{i}-\overline{\textbf{x}}_{j}})||_1}{||\textbf{w}^T\textbf{X}-\textbf{w}^T\overline{\textbf{x}}_{\textbf{I}}||_1}.
\end{equation}

We now derive a linear upper bound $g(x)$ of the right side of \eqref{Bhattacharyya0L1}.
Denote $h(x)=e^{-bx^2}$, $0\leq x\leq a$, $b>0$. It is easy to know that $h(x)$ is concave when $0\leq x \leq \frac{1}{\sqrt{2b}}$, and $h(x)$ is convex when $x \geq \frac{1}{\sqrt2b}$. Therefore, when $x\geq \frac{1}{\sqrt{2b}}$, the linear function passing through $(\frac{1}{\sqrt{2b}},h(\frac{1}{\sqrt{2b}}))$ and $(a,h(a))$ is the tightest linear upper bound of $h(x)$, e.g., $g(x)=-\frac{e^{-\frac{1}{2b}}-e^{-ba^2}}{a-\frac{1}{\sqrt{2b}}}x+\big(e^{-\frac{1}{2b}}-\frac{1}{\sqrt{2b}}\cdot \frac{e^{-ba^2}-e^{-\frac{1}{2b}}}{a-\frac{1}{\sqrt{2b}}}\big)$.
When $0\leq{x}\leq \frac{1}{\sqrt{2b}}$, it is obvious that there exists some constant $s>0$ such that $g(x)=-\frac{e^{-\frac{1}{2b}}-e^{-ba^2}}{a-\frac{1}{\sqrt{2b}}}x+\big(e^{-\frac{1}{2b}}-\frac{1}{\sqrt{2b}}\cdot \frac{e^{-ba^2}-e^{-\frac{1}{2b}}}{a-\frac{1}{\sqrt{2b}}}+s\big)$ is tangent to $h(x)$ and also a linear upper bound of $h(x)$.

In summary, if we define
\begin{equation}\label{}
\begin{split}
g(x)=-Ex+C,
\end{split}
\end{equation}
where $E=\frac{e^{-\frac{1}{2b}}-e^{-ba^2}}{a-\frac{1}{\sqrt{2b}}}$,
$C=e^{-\frac{1}{2b}}-\frac{1}{\sqrt{2b}}\cdot \frac{e^{-ba^2}-e^{-\frac{1}{2b}}}{a-\frac{1}{\sqrt{2b}}}+s$ if $0\leq{x}<\frac{1}{\sqrt{2b}}$ and $C=e^{-\frac{1}{2b}}-\frac{1}{\sqrt{2b}}\cdot \frac{e^{-ba^2}-e^{-\frac{1}{2b}}}{a-\frac{1}{\sqrt{2b}}}$ if $x\geq\frac{1}{\sqrt{2b}}$, $b=\frac{N}{8}$, then by combining \eqref{Bhattacharyya0L1} we have
\begin{equation}
\begin{split}
\epsilon_B
&\leq \sum\limits_{i<j}^c  \sqrt{P_iP_j} g\big(\frac{||\textbf{w}^T({\overline{\textbf{x}}_{i}-\overline{\textbf{x}}_{j}})||_1}{||\textbf{w}^T\textbf{X}-\textbf{w}^T\overline{\textbf{x}}_{\textbf{I}}||_1}\big)\\
&=\sum\limits_{i<j}^c\sqrt{P_iP_j}\big(-B\frac{||\textbf{w}^T({\overline{\textbf{x}}_{i}-\overline{\textbf{x}}_{j}})||_1}{||\textbf{w}^T\textbf{X}-\textbf{w}^T\overline{\textbf{x}}_{\textbf{I}}||_1}+C\big)\\
&\leq-B\sum\limits_{i<j}^c\sqrt{P_iP_j}\cdot ||\textbf{w}^T({\overline{\textbf{x}}_{i}-\overline{\textbf{x}}_{j}})||_1+B\Omega||\textbf{w}^T\textbf{X}-\textbf{w}^T\overline{\textbf{x}}_{\textbf{I}}||_1+C\sum\limits_{i<j}^c \sqrt{P_iP_j},\\
\end{split}
\end{equation}
where $\Omega= \frac{\sqrt{n}}{4}\sum\limits_{i<j}^c\sqrt{P_iP_j}||{\overline{\textbf{x}}_{i}-\overline{\textbf{x}}_{j}}||_1$ similar as in Proposition \ref{propL2}. \hfill$\square$

Therefore, by minimizing the upper bound \eqref{BhattacharyyaL1} of the Bhattacharyya error, we get the formulation of L1BLDA as
\begin{equation}
\begin{split}\label{L1BLDA}
\underset{\textbf{W}}{\min}~~&-\frac{1}{N}\sum_{i<j}\sqrt{N_iN_j}||\textbf{W}^T(\overline{\textbf{x}}_{i}-\overline{\textbf{x}}_{j})||_1+\Omega\sum_{i=1}^{c}\sum_{s=1}^{N_i}||\textbf{W}^T(\textbf{x}_s^i-\overline{\textbf{x}}_i)||_1\\
\hbox{s.t.\ }& \textbf{W}^T\textbf{W}=I,
\end{split}
\end{equation}
where $\textbf{W}\in\mathbb{R}^{n\times d}$, $d\leq n$.

Problem \eqref{L1BLDA} expresses the similar ideal of L2BLDA, but with the L2-norm terms in L2BLDA replaced with the L1-norm ones, and with different weighting constaints. Therefore,
minimizing \eqref{L1BLDA} makes the sum of the distances in the projected space between the centroid of the $i$-th class and the centroid not in the $i$-th class under the L1-norm measure as large as possible, also ensures each sample be close to its own class centroid.
As in L2BLDA, the coefficients in the first term of L1BLDA weight the distances between different classes, and $\Omega$ weights the between-class and the within-class scatters.
The orthogonormal constraint $\textbf{W}^T\textbf{W}=\textbf{I}$ again guarantees the minimum redundancy in the projected space.

\subsection{L1BLDA algorithm}
We now give the solving algorithm of L1BLDA. As we see, the non-smoothness and non-convexity of the objective of L1BLDA \eqref{L1BLDA} together with the orthonormal constraint make L1BLDA difficult to solve by traditional optimal techniques. Here we give an ADMM algorithm to solve it. To apply ADMM, we first transfer \eqref{L1BLDA} to its following ADMM form

\begin{equation}
\begin{split}\label{CCLDA2}
\underset{\textbf{W},\,\textbf{B}_{ij}\,\textbf{Z}_{is},\,\textbf{D}}{\min}&~~-\sum_{i<j}||\textbf{B}_{ij}||_1+\Omega\sum_{i=1}^{c}\sum_{s=1}^{N_i}||\textbf{Z}_{is}||_1
+\ell(\textbf{W})\\
\hbox{s.t.\ }&~~\frac{1}{N}\sqrt{N_iN_j}\textbf{W}^T(\overline{\textbf{x}}_i-\overline{\textbf{x}}_j)-\textbf{B}_{ij}=0,~~i<j,\\
&~~\textbf{W}^T(\textbf{x}_s^i-\overline{\textbf{x}}_i)-\textbf{Z}_{is}=0,\\
&~~\textbf{D}-\textbf{W}=0,\\
&~~i=1,\ldots,c,~~j=1,\ldots,N_i,
\end{split}
\end{equation}
where $\textbf{B}_{ij},\,\textbf{Z}_{is}\in\mathbb{R}^{d}$, $\textbf{W}, \textbf{D}\in \mathbb{R}^{n\times d}$, and $\ell(\textbf{W})=\begin{cases}0,~~&\textbf{W}^T\textbf{W}=I\\+\infty,&otherwise.\end{cases}$

Then the augmented Lagrangian is given by
\begin{eqnarray}
\begin{array}{cl}\label{augLag}
&L_\rho(\textbf{W},\textbf{B}_{ij},\textbf{Z}_{is},\textbf{D};\textbf{u}_{ij},\textbf{v}_{ij},\textbf{Q})\\
=&-\sum\limits_{i<j}^c||\textbf{B}_{ij}||_1+\Omega\sum\limits_{i=1}^{c}\sum\limits_{j=1}^{N_i}||\textbf{Z}_{is}||_1+\ell(\textbf{W})\\
&+\sum\limits_{i<j}^c\textbf{u}_{ij}^T(\frac{1}{N}\sqrt{N_iN_j}\textbf{W}^T(\overline{\textbf{x}}_i-\overline{\textbf{x}}_j)-\textbf{B}_{ij})\\
&+\sum\limits_{i=1}^{c}\sum\limits_{j=1}^{N_i}\textbf{v}_{ij}^T(\textbf{W}^T(\textbf{x}_s^i-\overline{\textbf{x}}_i)-\textbf{Z}_{is})+\textbf{Q}^T(\textbf{D}-\textbf{W})\\
&+\frac{\rho}{2}\sum\limits_{i<j}^c||\frac{1}{N}\sqrt{N_iN_j}\textbf{W}^T(\overline{\textbf{x}}_i-\overline{\textbf{x}}_j)-\textbf{B}_{ij}||_2^2\\
&+\frac{\rho}{2}\sum\limits_{i=1}^{c}\sum\limits_{j=1}^{N_i}||\textbf{W}^T(\textbf{x}_s^i-\overline{\textbf{x}}_i)-\textbf{Z}_{is}||_2^2+\frac{\rho}{2}||\textbf{D}-\textbf{W}||_F^2,
\end{array}
\end{eqnarray}
where $\textbf{u}_{ij}$, $\textbf{v}_{ij}\in\mathbb{R}^{d}$, $\textbf{Q}\in\mathbb{R}^{n\times d}$ are dual variables for $i=1,\ldots,c,~j=1,\ldots,N_i$, and $\rho>0$ is the penalty parameter. $<\cdot>$ denotes the inner product, where for two matrices $\textbf{P}=(p_{ij})$ and $\textbf{H}=(h_{ij})$ of the same size, their inner product is defined as $<\textbf{P},\textbf{H}>=\sum p_{ij}h_{ij}$.

By letting $\boldsymbol{\alpha}_{ij}=\textbf{u}_{ij}/\rho$, $\boldsymbol{\beta}_{ij}=\textbf{v}_{ij}/\rho$ and $\boldsymbol{\Gamma}=\textbf{Q}/\rho$ for $i=1,\ldots,c,~j=1,\ldots,N_i$, the scaled form Lagrangian of \eqref{augLag} can be formed as (without considering terms only contain $\boldsymbol{\alpha}_{ij}$, $\boldsymbol{\beta}_{ij}$ or $\boldsymbol{\Gamma}$)


\begin{eqnarray}
\begin{array}{cl}\label{augLagScaled}
&L_\rho(\textbf{W},\textbf{B}_{ij},\textbf{Z}_{is},\textbf{D};\boldsymbol{\alpha}_{ij},\boldsymbol{\beta}_{ij},\boldsymbol{\Gamma})\\
=&-\sum\limits_{i<j}^c||\textbf{B}_{ij}||_1+\Omega\sum\limits_{i=1}^{c}\sum\limits_{j=1}^{N_i}||\textbf{Z}_{is}||_1+\ell(\textbf{W})\\
&+\frac{\rho}{2}\sum\limits_{i<j}^c||\frac{1}{N}\sqrt{N_iN_j}\textbf{W}^T(\overline{\textbf{x}}_i-\overline{\textbf{x}})-\textbf{B}_{ij}+\boldsymbol{\alpha}_{ij}||_2^2\\
&+\frac{\rho}{2}\sum\limits_{i=1}^{c}\sum\limits_{j=1}^{N_i}||\textbf{W}^T(\textbf{x}_s^i-\overline{\textbf{x}}_i)-\textbf{Z}_{is}+\boldsymbol{\beta}_{ij}||_2^2\\
&+\frac{\rho}{2}||\textbf{D}-\textbf{W}+\boldsymbol{\Gamma}||_F^2.
\end{array}
\end{eqnarray}
Then the ADMM algorithm for problem \eqref{CCLDA2} can be presented as Algorithm 1.
For step (a) in Algorithm 1, we need to solve
\begin{equation}\label{Dk0}
\begin{split}
\textbf{W}^{k+1}=&\underset{\textbf{W}}{\arg\min}~L_\rho(\textbf{W},\textbf{B}_{ij}^k,\textbf{Z}_{is}^k,\textbf{D}^k;\boldsymbol{\alpha}_{ij}^k,\boldsymbol{\beta}_{ij}^k,\boldsymbol{\Gamma}^k)\\
=&\underset{\textbf{W}}{\arg\min}~\ell(\textbf{W})+\frac{\rho}{2}\sum\limits_{i<j}^c||\frac{1}{N}\sqrt{N_iN_j}\textbf{W}^T(\overline{\textbf{x}}_i-\overline{\textbf{x}}_j)-\textbf{B}_{ij}^k+\boldsymbol{\alpha}_{ij}^k||_2^2\\
&+\frac{\rho}{2}\sum\limits_{i=1}^{c}\sum\limits_{j=1}^{N_i}||\textbf{W}^T(\textbf{x}_s^i-\overline{\textbf{x}}_i)-\textbf{Z}_{is}^k+\boldsymbol{\beta}_{ij}^k||_2^2\\
&+\frac{\rho}{2}||\textbf{D}^k-\textbf{W}+\boldsymbol{\Gamma}^k||_2^2\\
=&\underset{\textbf{W}}{\arg\min}~\ell(\textbf{W})+\frac{\rho}{2}
tr(\textbf{W}^T(\sum\limits_{i<j}^c\frac{N_iN_j}{N^2}(\overline{\textbf{x}}_i-\overline{\textbf{x}}_j)(\overline{\textbf{x}}_i-\overline{\textbf{x}}_j)^T)\textbf{W})\\
&+\frac{\rho}{2}tr(\textbf{W}^T\sum\limits_{i=1}^{c}\sum\limits_{j=1}^{N_i}(\textbf{x}_s^i-\overline{\textbf{x}}_i)(\textbf{x}_s^i-\overline{\textbf{x}}_i)^T)\textbf{W})\\
&+\frac{\rho}{2}tr(\textbf{W}^T\textbf{W})-\rho\big[ tr(\sum\limits_{i<j}^c\frac{1}{N}\sqrt{N_iN_j}(\overline{\textbf{x}}_i-\overline{\textbf{x}}_j)^T\textbf{W}(\textbf{B}_{ij}^k-\boldsymbol{\alpha}_{ij}^k))\\
&+\rho\cdot tr(\sum\limits_{i=1}^{c}\sum\limits_{j=1}^{N_i}(\textbf{x}_s^i-\overline{\textbf{x}}_i)^T\textbf{W}(\textbf{Z}_{is}^k-\boldsymbol{\beta}_{ij}^k))\\
&+\rho\cdot tr((\textbf{D}^{k}+\boldsymbol{\Gamma}^{k})^T \textbf{W})\big]\\
\end{split}
\end{equation}
\begin{equation*}\label{Dk0c}
\begin{split}
=&\underset{\textbf{W}}{\arg\min}~\ell(\textbf{W})+\frac{\rho}{2}
tr[\textbf{W}^T(\sum\limits_{i<j}^c\frac{N_iN_j}{N^2}(\overline{\textbf{x}}_i-\overline{\textbf{x}}_j)(\overline{\textbf{x}}_i-\overline{\textbf{x}}_j)^T\\
&+\sum\limits_{i=1}^{c}\sum\limits_{j=1}^{N_i}(\textbf{x}_s^i-\overline{\textbf{x}}_i)(\textbf{x}_s^i-\overline{\textbf{x}}_i)^T+\textbf{I})\textbf{W}]\\
&-\rho\cdot tr[(\sum\limits_{i<j}^c\frac{1}{N}\sqrt{N_iN_j}(\textbf{B}_{ij}^k-\boldsymbol{\alpha}_{ij}^k)(\overline{\textbf{x}}_i-\overline{\textbf{x}}_j)^T\\
&+\sum\limits_{i=1}^{c}\sum\limits_{j=1}^{N_i}(\textbf{Z}_{is}^k-\boldsymbol{\beta}_{ij}^k)(\textbf{x}_s^i-\overline{\textbf{x}}_i)^T+(\textbf{D}^{k}+\boldsymbol{\Gamma}^{k})^T)\textbf{W}]\\
=&\underset{\textbf{W}}{\arg\min}~\ell(\textbf{W})+\frac{\rho}{2}
tr(\textbf{W}^TG\textbf{W})+\rho\cdot tr((A^k)^T\textbf{W}),\\
=&\underset{\textbf{W}}{\arg\min}~\frac{\rho}{2}
tr(\textbf{W}^TG\textbf{W})+\rho\cdot tr((A^k)^T\textbf{W})\\
&\hbox{s.t.\ }\textbf{W}^T\textbf{W}=I,
\end{split}
\end{equation*}
where $\textbf{G}=\sum\limits_{i<j}^c\frac{N_iN_j}{N^2}(\overline{\textbf{x}}_i-\overline{\textbf{x}}_j)(\overline{\textbf{x}}_i-\overline{\textbf{x}}_j)^T+\sum\limits_{i=1}^{c}\sum\limits_{j=1}^{N_i}(\textbf{x}_s^i-\overline{\textbf{x}}_i)(\textbf{x}_s^i-\overline{\textbf{x}}_i)^T+\textbf{I}\in\mathbb{R}^{n\times n}$, and $(\textbf{A}^k)^T=\sum\limits_{i<j}^c\frac{1}{N}\sqrt{N_iN_j}(\textbf{B}_{ij}^k-\boldsymbol{\alpha}_{ij}^k)(\overline{\textbf{x}}_i-\overline{\textbf{x}}_j)^T+\sum\limits_{i=1}^{c}\sum\limits_{j=1}^{N_i}(\textbf{Z}_{is}^k-\boldsymbol{\beta}_{ij}^k)(\textbf{x}_s^i-\overline{\textbf{x}}_i)^T+(\textbf{D}^{k}+\boldsymbol{\Gamma}^{k})^T\in\mathbb{R}^{d\times n}$. The above problem is equivalent to
\begin{equation}\label{Dk1}
\begin{split}
&\underset{\textbf{W}}{\arg\min}~\frac{\rho}{2}
[tr(\textbf{W}^TG\textbf{W})+2\cdot tr((\textbf{A}^k)^T\textbf{W})]\\
&\hbox{s.t.\ }\textbf{W}^T\textbf{W}=I,
\end{split}
\end{equation}

\begin{algorithm}\label{ADMM}
\caption{~Scaled ADMM algorithm for problem \eqref{L2BLDA}.}
\begin{tabular}{ll}
\textbf{Input:} Data set $T=\{({x}_{1},y_{1}),...,({x}_{m},y_{m})\}$; the positive tolerances $\epsilon^{pri}$ \\and $\epsilon^{dual}$.
Set the iteration number $k=0$ and initialize $\textbf{D}^0\in \mathbb{R}^{n\times d},\textbf{B}_{ij}^0\in \mathbb{R}^{d}$,\\$\textbf{Z}_{is}^0\in \mathbb{R}^{d}$
and $\boldsymbol{\alpha}_{ij}^0,\boldsymbol{\beta}_{ij}^0\in \mathbb{R}^{d}$, $\boldsymbol{\Gamma}^0\in \mathbb{R}^{n\times d}$, $i=1,\ldots,c$, $j=1,\ldots,N_i$; maximum \\iteration number \textit{ItMax}.\\
\textbf{Process:}\\
{\bf while} $k<ItMax$\\
\quad(a)
$\textbf{W}^{k+1}=\underset{\textbf{W}}{\arg\min}L_\rho(\textbf{W},\textbf{B}_{ij}^k,\textbf{Z}_{is}^k,\textbf{D}^k;\boldsymbol{\alpha}_{ij}^k,\boldsymbol{\beta}_{ij}^k,\boldsymbol{\Gamma}^k)$;\\
\quad(b)
$\textbf{B}_{ij}^{k+1}=\underset{\textbf{B}_{ij}}{\arg\min}L_\rho(\textbf{W}^{k+1},\textbf{B}_{ij},\textbf{Z}_{is}^k,\textbf{D}^{k};\boldsymbol{\alpha}_{ij}^k,\boldsymbol{\beta}_{ij}^k,\boldsymbol{\Gamma}^k)$;\\
\quad(c)
$\textbf{Z}_{is}^{k+1}=\underset{\textbf{Z}_{is}}{\arg\min}L_\rho(\textbf{W}^{k+1},\textbf{B}_{ij}^{k+1},\textbf{Z}_{is},\textbf{D}^{k};\boldsymbol{\alpha}_{ij}^k,\boldsymbol{\beta}_{ij}^k,\boldsymbol{\Gamma}^k)$;\\
\quad(d)
$\textbf{D}^{k+1}=\underset{\textbf{Z}_{is}}{\arg\min}L_\rho(\textbf{W}^{k+1},\textbf{B}_{ij}^{k+1},\textbf{Z}_{is}^{k+1},\textbf{D};\boldsymbol{\alpha}_{ij}^k,\boldsymbol{\beta}_{ij}^k,\boldsymbol{\Gamma}^k)$;\\
\quad(e)
$\boldsymbol{\alpha}_{ij}^{k+1}=\boldsymbol{\alpha}_{ij}^{k}+(\frac{1}{N}\sqrt{N_iN_j}(\textbf{W}^{k+1})^T(\overline{\textbf{x}}_i-\overline{\textbf{x}}_j)-\textbf{B}_{ij}^{k+1})$;\\
\quad(f)
$\boldsymbol{\beta}_{ij}^{k+1}=\boldsymbol{\beta}_{ij}^{k}+((\textbf{W}^{k+1})^T(\textbf{x}_s^i-\overline{\textbf{x}}_i)-\textbf{Z}_{is}^{k+1})$;\\
\quad(g)
$\boldsymbol{\Gamma}^{k+1}=\boldsymbol{\Gamma}^{k}+(\textbf{D}^{k+1}-\textbf{W}^{k+1})$\\

\textbf{Until} \\ $||r^{k+1}||=\underset{i,j}{\max}\{||\frac{1}{N}\sqrt{N_iN_j}(\textbf{W}^{k+1})^T(\overline{\textbf{x}}_i-\overline{\textbf{x}}_j)-\textbf{B}_{ij}^{k+1}||_2$,\\$||(\textbf{W}^{k+1})^T(\textbf{x}_s^i-\overline{\textbf{x}}_i)-\textbf{Z}_{is}^{k+1}||_2$, $||\textbf{W}^{k+1}-\textbf{D}^{k+1}||_2\}\leq\epsilon^{pri}$\\
and\\
$||s^{k+1}||=\underset{i,j}\max\{||\rho (\overline{\textbf{x}}_i-\overline{\textbf{x}}_j)(\textbf{B}_{ij}^{k+1}-\textbf{B}_{ij}^{k})'||_2$, \\$||\rho (\textbf{x}_s^i-\overline{\textbf{x}}_i)(\textbf{Z}_{is}^{k+1}-\textbf{Z}_{is}^{k})'||_2$, $||\rho(\textbf{D}^{k+1}-\textbf{D}^k)||_F\}\leq\epsilon^{dual}$.\\

\textbf{Output:} $\textbf{W}^*=\textbf{W}^{k+1}$. \\
\end{tabular}
\end{algorithm}

From Algorithm 1, we see that we need to solve optimization problems in steps (a)-(d). In the following, we will give specific solutions to these four types of problems.

Now we solve problem \eqref{Dk1} by two cases.\\
\textbf{Case 1:} $d=n$.
Since $\textbf{G}$ is positive definite, we can further write $\textbf{G}=\textbf{G}_0(\textbf{G}_0)^T$ by Cholesky decomposition, where $\textbf{G}_0$ is an invertible lower triangular matrix. Then problem \eqref{Dk0} is equivalent to

\begin{equation}\label{Dk2}
\begin{split}
\underset{\textbf{W}}{\arg\min}&~~\frac{\rho}{2}||\textbf{G}_0^T\textbf{W}-(\textbf{G}_0)^{-1}\textbf{A}^k||_F^2\\
\hbox{s.t.\ }&~~\textbf{W}^T\textbf{W}=\textbf{I}.
\end{split}
\end{equation}

In this situation, problem \eqref{Dk2} is a balanced Procrustes problem \cite{Procrustes}, and can be solved as
$\textbf{W}^{k+1}=\textbf{P}_1^k(\textbf{P}_2^k)^T$, where $\textbf{P}_1^k$ and $\textbf{P}_2^k$ are orthogonal matrices from the SVD
\begin{equation*}\label{}
\textbf{A}^k=\textbf{P}_1^k\boldsymbol{\Sigma}^k\textbf{P}_2^k.
\end{equation*}\\
\textbf{Case 2:} $d<n$. In this situation, problem \eqref{Dk1} is an unbalanced Procrustes problem \cite{Mulaik72}, and there is no analytic solution. We here adopt a recently proposed convergent algorithm studied in \cite{NieZhangLi17} to solve \eqref{Dk1}. Specifically, we use Algorithm 2.

\begin{algorithm}\label{}
\caption{~Algorithm for problem \eqref{Dk1} when $d<n$.}
\begin{tabular}{ll}
(a) Compute the dominant eigenvalue $a$ of $\textbf{G}$.\\
(b) Randomly initialize $\textbf{W}\in \mathbb{R}^{n\times d}$ such that $\textbf{W}^T\textbf{W}=\textbf{I}$.\\
(c) Update $\textbf{M}\leftarrow 2(a\textbf{I}-\textbf{G})\textbf{W}-2\textbf{A}^k$.\\
(d) Calculate $\textbf{W}$ by solving the following problem:\\
\quad \quad $\underset{\textbf{W}^T\textbf{W}=\textbf{I}}{\max} tr(\textbf{W}^T\textbf{M}).$\\
(e) Iteratively perform the above steps (c) and (d) until \\
\quad convergence.
\end{tabular}
\end{algorithm}

For step (b) in Algorithm 1, we need to solve
\begin{equation}\label{}
\begin{split}
\textbf{B}_{ij}^{k+1}=&\underset{\textbf{B}_{ij}}{\arg\min}~L_\rho(\textbf{W}^{k+1},\textbf{B}_{ij},\textbf{Z}_{is}^k,\textbf{D}^k;\boldsymbol{\alpha}_{ij}^k,\boldsymbol{\beta}_{ij}^k,\boldsymbol{\Gamma}^k)\\
=&\underset{\textbf{B}_{ij}}{\arg\min}~-\sum\limits_{i<j}^c||\textbf{B}_{ij}||_1\\
&+\frac{\rho}{2}\sum\limits_{i<j}^c||\frac{1}{N}\sqrt{N_iN_j}(\textbf{W}^k)^T(\overline{\textbf{x}}_i-\overline{\textbf{x}}_j)-\textbf{B}_{ij}+\boldsymbol{\alpha}_{ij}^k||_2^2.\\
\end{split}
\end{equation}
By direct computation, its solution can be given as
$$\textbf{B}_{ij}^{k+1}=\begin{cases}&\frac{1}{N}\sqrt{N_iN_j}(\textbf{W}^k)^T(\overline{\textbf{x}}_i-\overline{\textbf{x}}_j)+\boldsymbol{\alpha}_{ij}^k+1/\rho,\\&~~~\textrm{if~} \frac{1}{N}\sqrt{N_iN_j}(\textbf{W}^k)^T(\overline{\textbf{x}}_i-\overline{\textbf{x}}_j)+\boldsymbol{\alpha}_{ij}^k\geq0;\\
&\frac{1}{N}\sqrt{N_iN_j}(\textbf{W}^k)^T(\overline{\textbf{x}}_i-\overline{\textbf{x}}_j)+\boldsymbol{\alpha}_{ij}^k-1/\rho,\\&~~~\textrm{if~} \frac{1}{N}\sqrt{N_iN_j}(\textbf{W}^k)^T(\overline{\textbf{x}}_i-\overline{\textbf{x}}_j)+\boldsymbol{\alpha}_{ij}^k<0,\end{cases}$$ for $i=1,2,\ldots,c$.

For step (c) in Algorithm 1, we need to solve
\begin{equation}\label{}
\begin{split}
\textbf{Z}_{is}^{k+1}=&\underset{\textbf{Z}_{is}}{\arg\min}L_\rho(\textbf{W}^{k+1},\textbf{B}_{ij}^{k+1},\textbf{Z}_{is},\textbf{D}^k;\boldsymbol{\alpha}_{ij}^k,\boldsymbol{\beta}_{ij}^k,\boldsymbol{\Gamma}^k)\\
=&\underset{\textbf{Z}_{is}}{\arg\min}~\Omega\sum\limits_{i=1}^{c}\sum\limits_{j=1}^{N_i}||\textbf{Z}_{is}||_1
+\frac{\rho}{2}\sum\limits_{i=1}^{c}\sum\limits_{j=1}^{N_i}||(\textbf{W}^{k+1})^T(\textbf{x}_s^i-\overline{\textbf{x}}_i)-\textbf{Z}_{is}+\boldsymbol{\beta}_{ij}^{k}||_2^2.\\
\end{split}
\end{equation}
Its solution can be given through soft thresholding function:
$$\textbf{Z}_{is}^{k+1}=\boldsymbol{\Phi}_{\Omega/\rho}[(\textbf{W}^{k+1})^T(\textbf{x}_s^i-\overline{\textbf{x}}_i)+\boldsymbol{\beta}_{ij}^{k}]$$ for $i=1,2,\ldots,c$ and $j=1,2,\ldots,N_i$, where $\boldsymbol{\Phi}_\kappa(a)=\begin{cases}a-\kappa,~~~a>\kappa\\0,~~~~~~~~|a|\leq \kappa\\a+\kappa,~~~a<-\kappa\end{cases}$.

For step (d) in Algorithm 1, we need to solve
\begin{equation}\label{}
\begin{split}
\textbf{D}^{k+1}=&\underset{\textbf{D}}{\arg\min~}L_\rho(\textbf{W}^{k+1},\textbf{B}_{ij}^{k+1},\textbf{Z}_{is}^{k+1},\textbf{D};\boldsymbol{\alpha}_{ij}^k,\boldsymbol{\beta}_{ij}^k,\boldsymbol{\Gamma}^k)\\
=&\underset{\textbf{D}}{\arg\min~}\frac{\rho}{2}||\textbf{D}-\textbf{W}^{k+1}+\boldsymbol{\Gamma}^{k}||_F^2,\\
\end{split}
\end{equation}
whose solution is componentwisely given by $$\textbf{D}^{k+1}=\textbf{W}^{k+1}-\boldsymbol{\Gamma}^{k}.$$

\section{Experiments}\label{secExp}

In this section, we perform experiments to test the proposed methods on some UCI benchmark data sets and two handwritten digit databases. Several related dimensionality reduction methods, including PCA \cite{TurkPentland91}, PCA-L1\cite{Kwak08}, LDA \cite{Fukunaga90}, LDA-L1 \cite{ZhongZhang13}
are used for comparison.
The learning rate parameters for LDA-L1 is chosen from the set $\{10^{-4}, 10^{-3}, \ldots, 10^{1}\}$.
To test the discriminant ability of various methods, the nearest neighbor classification accuracy ($\%$) in the projected space is used as an indicator, where the projected space is obtained by applying each dimensionality reduction method on the training data.
All the methods are carried out on a PC with P4 2 GHz CPU and 2 GB RAM memory by Matlab 2017b.

\subsection{The UCI data sets}
\begin{figure}[htpb]
\begin{center}{
{\resizebox*{9cm}{!}
{\includegraphics{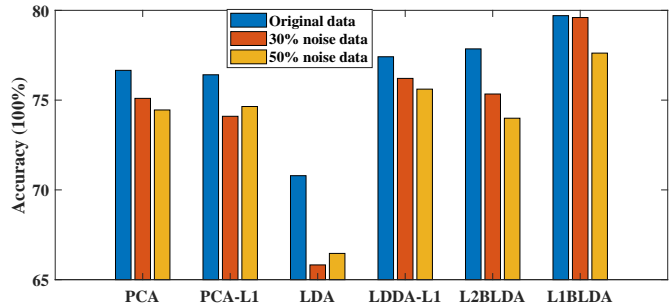}}}\hspace{5pt}
\caption{The mean accuracies for various methods on the UCI data sets.}
\label{UCImeanACC}}
\end{center}
\end{figure}

We first apply the proposed methods on 21 benchmark data sets. All the data sets are normalized to the interval $[0,1]$, and their information is listed in Table \ref{TableUCIInfo}. Random 70\% percent data is used for training, and the rest data forms the test set.
To test the robustness of our L1BLDA, we also add random Gaussian noise to each training data. In specific, 30\% and 50\% percent features are added with random Gaussian noise of mean zero and variance 0.1, respectively. The classification results on the original data sets and the noise data sets are listed in Table \ref{TableUCIori}, Table \ref{TableUCI30} and Table \ref{TableUCI50}, respectively.
From the tables, we see no matter which situation, L1-norm based methods generally perform better than the L2-norm ones, and our L1BLDA performs the best comparing to the other methods. When the noise is added, the performance for all the methods degenerates on almost all the data sets. However, our L1BLDA is less affected comparing to the other methods. To see the results more clearer, we also depict the mean accuracies for each method over all the data sets in Figure \ref{UCImeanACC}, and list their average ranks in Table \ref{TableUCIRank}. One can see that our L1BLDA outperforms the others.
\begin{table*}[htbp]
\begin{center}
\renewcommand\tabcolsep{1pt}
\caption{UCI data sets information.}
\resizebox{5.5in}{!}
{
\begin{tabular}{lccccclccccc}
\toprule
{\multirow{1}{*}{Data set}}& \vline &Sample no.&Feature no.&Class no.& \vline &{\multirow{1}{*}{Data set}}& \vline &Sample no.&Feature no.&Class no.\\
\midrule
Australian& \vline&690&14&2&\vline&Iris&\vline&150&4&3\\
BUPA& \vline&345&6&2&\vline&Monks3&\vline&432&6&2\\
Car& \vline&1782&6&4&\vline&Musk1&\vline&476&166&2\\
Credit& \vline&690&15&2&\vline&Libras&\vline&360&90&15\\
Diabetics& \vline&768&8&2&\vline&Sonar&\vline&208&60&2\\
Echocardiogram& \vline&131&10&2&\vline&Spect&\vline&267&44&2\\
Ecoli& \vline&336&7&8&\vline&TicTacToe&\vline&958&27&2\\
German& \vline&1000&20&2&\vline&Titanic&\vline&2201&3&2\\
Haberman& \vline&306&3&2&\vline&Waveform&\vline&5000&21&2\\
Hourse& \vline&300&2&2&\vline&WPBC&\vline&198&34&2\\
House$\_$votes& \vline&435&16&2&\vline&&\vline\\
\bottomrule
\end{tabular}
}
\label{TableUCIInfo}
\end{center}
\end{table*}

\begin{table*}[htbp]
\begin{center}
\caption{Classification results on original UCI data sets.}
\resizebox{5.5in}{!}
{
\begin{tabular}{lccccccccccc}
\toprule
{\multirow{2}{*}{Data set}}& \vline &PCA&PCA-L1& LDA &LDA-L1&L2BLDA&L1BLDA\\
& \vline& Acc (Dim)& Acc (Dim)&Acc (Dim)& Acc (Dim)& Acc (Dim)& Acc (Dim)\\
\midrule
Australian& \vline&81.16 (7)&80.19 (6)&80.19 (1)&82.31 (8)&82.31 (2)&\textbf{83.57 (3)}\\
BUPA& \vline&60.19 (6)&61.17 (2)&56.31 (1)&63.11 (6)&65.05 (3)&\textbf{70.87 (4)}\\
Car& \vline&93.63 (3)&76.25 (6)&50.00 (3)&75.87 (5)&\textbf{93.82 (5)}&92.28 (5)\\
Credit& \vline&81.64 (5)&\textbf{82.61 (3)}&80.68 (1)&\textbf{82.61 (8)}&81.16 (6)&\textbf{82.61 (5)}\\
Diabetics& \vline&71.74 (4)&70.87 (8)&68.70 (1)&\textbf{73.48 (7)}&70.87 (8)&72.17 (5)\\
Echocardiogram& \vline&\textbf{87.18 (9)}&\textbf{87.18 (8)}&\textbf{87.18 (1)}&\textbf{87.18 (10)}&\textbf{87.18 (9)}& \textbf{87.18 (9)}\\
Ecoli& \vline&78.22 (7)&\textbf{80.20 (4)}&77.23 (7)&77.23 (7)&78.22 (5)&\textbf{80.20 (6)}\\
German& \vline&73.67 (18)&73.67 (17)&69.33 (1)&74.33 (15)&73.67 (18)&\textbf{74.67 (3)}\\
Haberman& \vline&71.43 (3)&71.43 (3)&51.65 (1)&68.13 (2)&73.63 (2)&\textbf{74.73 (2)}\\
Hourse& \vline&\textbf{84.44 (11)}&80.00 (11)&64.44 (1)&82.22 (17)&83.33 (11)& \textbf{84.44 (13)}\\
House$\_$votes& \vline&88.46 (18)&87.69 (16)&\textbf{91.54 (1)}&90.77 (7)&90.77 (16)& \textbf{91.54 (16)}\\
Iris& \vline&\textbf{100 (1)}&\textbf{100 (3)}&\textbf{100 (2)}&\textbf{100 (3)}&\textbf{100 (4)}&\textbf{100 (3)}\\
Monks3&\vline&63.08 (3)&60.77 (3)&60.00 (1)&70.00 (1)&56.15 (5)&\textbf{73.85 (1)}\\
Musk1& \vline&63.08 (41)&\textbf{83.92 (44)}&76.92 (1)&\textbf{83.92 (45)}&\textbf{83.92 (45)}&\textbf{83.92 (44)}\\
Libras&\vline&\textbf{52.38 (15)}&\textbf{52.38 (15)}&44.76 (14)&\textbf{52.38 (11)}&\textbf{52.38 (15)}& \textbf{52.38 (16)}\\
Sonar& \vline&56.45 (9)&56.45 (9)&46.77 (1)&\textbf{62.90 (7)}&56.45 (9)&\textbf{62.90 (5)}\\
Spect&\vline&\textbf{83.75 (5)}&78.75 (4)&71.25 (1)&78.75 (2)&81.25 (5)&\textbf{83.75 (6)}\\
TicTacToe&\vline&97.57 (14)&96.88 (15)&94.44 (1)&95.83 (15)&\textbf{99.31 (14)}& 93.75 (15)\\
Titanic& \vline&67.73 (1)&\textbf{70.30 (1)}&67.73 (1)&\textbf{70.30 (2)}&67.73 (1)&\textbf{70.30 (1)}\\
Waveform& \vline&86.23 (3)&86.13 (3)&81.40 (1)&\textbf{86.60 (4)}&86.53 (3)&85.80 (9)\\
WPBC& \vline&67.80 (1)&67.80 (1)&66.10 (1)&67.80 (2)&71.19 (7)&\textbf{72.88 (1)}\\
\bottomrule
\end{tabular}
}
\label{TableUCIori}
\end{center}
\end{table*}

\begin{table*}[htbp]
\begin{center}
\caption{Classification results on the UCI data sets with 30\% features added with Gaussian noise.}
\resizebox{5.5in}{!}
{
\begin{tabular}{lccccccccccc}
\toprule
{\multirow{2}{*}{Data set}}& \vline &PCA&PCA-L1& LDA &LDA-L1&L2BLDA&L1BLDA\\
& \vline& Acc  (Dim)& Acc  (Dim)&Acc  (Dim)& Acc  (Dim)& Acc  (Dim)& Acc (Dim)\\
\midrule
Australian& \vline&81.64 (9)&81.64 (2)&78.74 (1)&\textbf{84.54 (7)}&81.64 (2)&\textbf{84.54 (7)}\\
BUPA& \vline&56.31 (6)&57.28 (1)&57.28 (1)&60.19 (2)&56.31 (3)&\textbf{68.93 (1)}\\
Car& \vline&81.66 (4)&70.08 (6)&45.56 (3)&67.57 (2)&81.80 (3)&\textbf{84.56 (5)}\\
Credit& \vline&83.57 (15)&84.06 (9)&78.26 (1)&\textbf{86.47 (12)}&83.57 (15)&85.99 (7)\\
Diabetics& \vline&73.04 (8)&73.04 (8)&63.04 (1)&74.35 (3)&73.04 (8)&\textbf{75.65 (7)}\\
Echocardiogram& \vline&76.92 (1)&82.05 (6)&74.36 (1)&\textbf{84.62 (2)}&79.49 (8)& \textbf{84.62 (2)}\\
Ecoli& \vline&77.23 (4)&75.25 (7)&75.25 (5)&75.25 (5)&75.25 (6)&\textbf{78.22 (4)}\\
German& \vline&71.67 (8)&73.00 (8)&67.33 (1)&72.33 (18)&72.00 (5)&\textbf{73.33 (6)}\\
Haberman& \vline&\textbf{75.82 (3)}&\textbf{75.82 (3)}&61.54 (1)&72.53 (1)&\textbf{75.82 (3)}&\textbf{75.82 (3)}\\
Hourse& \vline&80.00 (9)&77.78 (8)&65.56 (1)&\textbf{85.56 (8)}&78.89 (10)&82.22 (8)\\
House$\_$votes& \vline&89.23 (11)&87.69 (16)&90.77 (1)&90.00 (10)&87.69 (16)& \textbf{92.31 (3)}\\
Iris& \vline&88.89(2)&91.11 (3)&\textbf{97.78 (2)}&93.33 (3)&93.33 (3)&\textbf{97.78 (2)}\\
Monks3&\vline&65.38 (2)&47.69 (1)&51.54 (5)&53.08 (1)&74.62 (2)&\textbf{76.92 (5)}\\
Musk1& \vline&79.72 (27)&78.32 (21)&65.73 (1)&79.72 (11)&79.72 (48)&\textbf{83.22 (12)}\\
Libras&\vline&60.00 (24)&57.14 (37)&39.05 (13)&\textbf{61.90 (34)}&57.14 (23)& 56.19 (24)\\
Sonar& \vline&45.16 (3)&50.00 (3)&54.84 (1)&54.84 (1)&43.55 (6)&\textbf{67.74 (1)}\\
Spect&\vline&75.00 (9)&\textbf{82.50 (18)}&60.00 (1)&81.25 (6)&73.75 (9)&80.00 (7)\\
TicTacToe&\vline&90.97 (6)&91.67 (18)&45.14 (1)&\textbf{92.71 (16)}&91.32 (7)& \textbf{92.71 (15)}\\
Titanic& \vline&\textbf{70.30 (1)}&\textbf{70.30 (1)}&\textbf{70.30 (1)}&\textbf{70.30 (1)}&\textbf{70.30 (1)}&\textbf{70.30 (1)}\\
Waveform& \vline&81.67 (8)&81.87 (12)&80.86 (1)&83.60 (17)&81.73 (12)&\textbf{84.27 (6)}\\
WPBC& \vline&72.88 (15)&67.80 (7)&59.32 (1)&\textbf{76.27 (25)}&71.19 (3)&\textbf{76.27 (4)}\\
\bottomrule
\end{tabular}
}
\label{TableUCI30}
\end{center}
\end{table*}

\begin{table*}[htbp]
\begin{center}
\caption{Classification results on the UCI data sets with 50\% features added with Gaussian noise.}
\resizebox{5.5in}{!}
{
\begin{tabular}{lccccccccccc}
\toprule
{\multirow{2}{*}{Data set}}& \vline &PCA&PCA-L1& LDA &LDA-L1&L2BLDA&L1BLDA\\
& \vline& Acc  (Dim)& Acc  (Dim)&Acc  (Dim)& Acc  (Dim)& Acc  (Dim)& Acc (Dim)\\
\midrule
Australian& \vline&82.61 (7)&83.57 (3)&78.74 (1)&83.09 (5)&83.09 (7)&\textbf{84.06 (6)}\\
BUPA& \vline&58.25 (2)&57.28 (4)&62.14 (1)&59.22 (2)&60.19 (2)& \textbf{67.96 (4)}\\
Car& \vline&78.19 (5)&70.08 (6)&44.21 (3)&68.73 (4)&77.03 (5)&\textbf{78.57 (4)}\\
Credit& \vline&79.71 (4)&76.81 (6)&78.26 (1)&\textbf{81.64 (6)}&76.81(15)& \textbf{81.64 (4)}\\
Diabetics& \vline&66.52 (8)&\textbf{69.57 (4)}&62.17 (1)&67.83 (6)&66.52 (8)&\textbf{69.57 (3)}\\
Echocardiogram& \vline&\textbf{84.62 (2)}&79.49 (4)&79.49 (1)&\textbf{84.62 (8})&82.50 (7)& \textbf{84.62 (3)}\\
Ecoli& \vline&\textbf{78.22 (7)}&\textbf{78.22 (7)}&75.25 (7)&73.27 (5)&\textbf{78.22 (7)}&\textbf{78.22 (7)}\\
German& \vline&70.33 (5)&70.67 (10)&68.33 (1)&73.67 (8)&69.33 (14)&\textbf{74.67 (3)}\\
Haberman& \vline&70.33 (3)&71.43 (3)&63.74 (1)&70.33 (3)&70.33 (3)&\textbf{74.73 (2)}\\
Hourse& \vline& 83.33(17)&82.22(23)&76.67(1)&82.22(26)& 84.44(15)&\textbf{85.56(18)}\\
House$\_$votes&\vline&90.00(7)&\textbf{92.31(12)}&91.54(1)&\textbf{92.31(15)}& 90.00(16)&\textbf{92.31(5)}\\
Iris& \vline&88.89 (1)&84.44 (2)&82.22 (1)&86.67 (3)&84.44 (4)&\textbf{91.11 (2)}\\
Monks3&\vline&51.54 (1)&65.38 (3)&48.46 (1)&\textbf{72.31 (2)}&49.23 (5)&70.77 (2)\\
Musk1& \vline&77.62 (35)&77.62 (129)&62.24 (1)&\textbf{78.33 (93)}&77.62 (35)& \textbf{78.33 (38)}\\
Libras&\vline&59.05 (48)&\textbf{61.90 (54)}&42.86 (3)&60.95 (60)&55.24 (39)&59.05 (29)\\
Sonar& \vline&51.61 (6)&56.45 (1)&48.39 (1)&\textbf{59.68 (3)}&53.23 (7)&\textbf{59.68 (7)}\\
Spect&\vline&77.50 (7)&78.75 (8)&70.00 (1)&80.00 (7)&78.75 (7)&\textbf{81.25 (8)}\\
TicTacToe&\vline&\textbf{94.44 (9)}&92.36 (15)&58.68 (1)&93.40 (16)&\textbf{94.44 (9)}& 89.93 (15)\\
Titanic& \vline&67.73 (1)&67.73 (1)&67.73 (1)&67.73 (1)&67.73 (1)&\textbf{70.30 (1)}\\
Waveform& \vline&76.73 (11)&76.67 (19)&75.27 (1)&77.27 (16)&76.73 (20)& \textbf{78.00 (19)}\\
WPBC& \vline&76.26 (8)&74.58 (7)&59.32 (1)&74.58 (27)&77.97 (9)&\textbf{79.66 (6)}\\
\bottomrule
\end{tabular}
}
\label{TableUCI50}
\end{center}
\end{table*}

\begin{table*}[htbp]
\begin{center}
\caption{Average ranks of various methods for the accuracies on the UCI data sets.}
\resizebox{4.5in}{!}
{
\begin{tabular}{lccccccccccc}
\toprule
{\multirow{1}{*}{Data set}}& \vline &PCA&PCA-L1& LDA &LDA-L1&L2BLDA&L1BLDA\\
\midrule

Australian& \vline&4.00 &3.50 &5.83 &2.50 &3.00 &1.17\\
BUPA&  \vline&5.17 &4.50 &3.83 &4.17 &2.33 &1.00 \\
Car&  \vline&2.33 &4.00 &6.00 &5.00 &2.00 &1.67\\
CMC&  \vline&2.33 &5.00 &5.67 &2.00 &4.17 &1.83\\
Credit&  \vline&3.83 &3.00 &5.83 &1.50 &5.00 &1.83\\
Diabetics&  \vline&3.83 &3.17 &5.83 &2.00 &4.67 &1.50\\
Echocardiogram&  \vline&3.50 &4.00 &5.00 &2.33 &3.83 &2.33\\
Ecoli&  \vline&2.67 &2.83 &5.00 &5.33 &3.50 &1.67\\
German& \vline&4.33 &3.00 &6.00 &2.33 &4.33 &1.00\\
Glass&  \vline&4.00 &2.50 &4.83 &3.00 &5.33 &1.33\\
Haberman&  \vline&3.33 &2.67 &6.00 &4.67 &2.83 &1.50\\
Heartstatlog&  \vline&4.50 &3.33 &3.67 &2.67 &5.83 &1.00\\
Hourse&  \vline&2.50 &4.83 &6.00 &3.17 &3.00 &1.50\\
House$\_$votes&  \vline&4.83 &4.50 &2.50 &2.83 &4.83 &1.50\\
Ionosphere&  \vline&3.33 &4.33 &5.33 &1.83 &5.00 &1.17\\
Iris&  \vline&3.83 &4.33 &3.67 &3.33 &3.83 &2.00\\
Monks3& \vline&3.33 &4.33 &5.33 &2.33 &4.33 &1.33\\
Musk1& \vline&4.00 &3.83 &4.83 &2.83 &2.83 &1.67\\
Libras& \vline&2.83 &2.50 &6.00 &2.00 &3.83 &3.83\\
Sonar&  \vline&4.67 &3.67 &4.83 &1.83 &4.67 &1.33\\
Spect& \vline&3.50 &3.00 &6.00 &2.83 &3.83 &1.83\\
TicTacToe& \vline&2.83 &3.33 &5.67 &2.83 &2.17 &4.17\\
Titanic&  \vline&1.83 &3.17 &2.83 &3.17 &2.83 &3.17\\
Waveform&  \vline&3.83 &4.00 &6.00 &1.67 &3.17 &2.33\\
WPBC&  \vline&3.50 &4.67 &6.00 &3.00 &2.67 &1.17\\
\hline
Average rank&  \vline&3.55&	3.66&5.19&2.94&3.50&1.88\\
\bottomrule
\end{tabular}
}
\label{TableUCIRank}
\end{center}
\end{table*}

\subsection{Handwritten digit databases}

In this subsection, the behaviors of various methods are investigated on two handwritten digit databases, including the MNIST database and the USPS database.

\subsubsection{The MNIST database}
\begin{figure}[htpb]
\begin{center}{
{\resizebox*{8.5cm}{!}
{\includegraphics{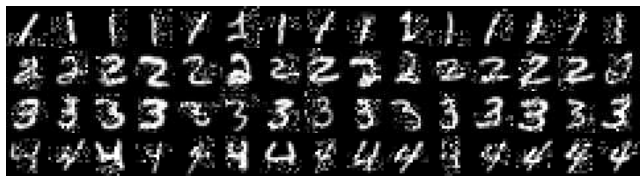}}}\hspace{5pt}
\caption{The contaminated samples from the MNIST database.}
\label{MNISTsample}}
\end{center}
\end{figure}

The MNIST database contains 70000 digit images with 10 classes of the size 28$\times$28. We up-sample the images to the size 16$\times$16, and further reshape them to vectors of the length 256. 30\% data from each class are randomly selected for training, while the rest data is used for testing. Further,
Gaussian noise with mean 0 and variance 0.05 is added on the training data, where the noise covers random 30\% rectangular area of each image. The contaminated digit images are displayed in Figure \ref{MNISTsample}.
All the methods are then applied on the original training data and the contaminated training data. We show the classification results in Table \ref{TableMNIST}.

\begin{table*}[htbp]
\begin{center}
\caption{Classification results on the MNIST database.}
\resizebox{5.5in}{!}
{
\begin{tabular}{lccccccccccc}
\toprule
{\multirow{2}{*}{Data}}& \vline &PCA&PCA-L1& LDA &LDA-L1&L2BLDA&L1BLDA\\
& \vline& Acc  (Dim)& Acc  (Dim)&Acc  (Dim)& Acc  (Dim)& Acc  (Dim)& Acc  (Dim)\\
\midrule
Original data& \vline& 96.79 (53)&\textbf{96.81 (66)}&86.24 (9)&96.50 (246)&\textbf{96.81 (56)}& 96.79 (57)\\
Noise data& \vline&71.90 (225)&72.27 (256)&77.57 (9)&91.41 (166)&92.28 (151)&\textbf{92.31 (156)}
 \\

\bottomrule
\end{tabular}
}
\label{TableMNIST}
\end{center}
\end{table*}

\begin{figure}[htpb]
\begin{center}{
\subfigure[Original data]{
\resizebox*{9.5cm}{!}
{\includegraphics{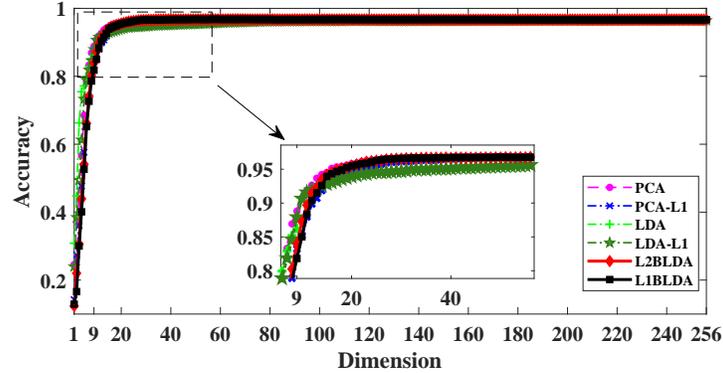}}}\hspace{5pt}
\subfigure[Noise data]{
\resizebox*{9.5cm}{!}
{\includegraphics{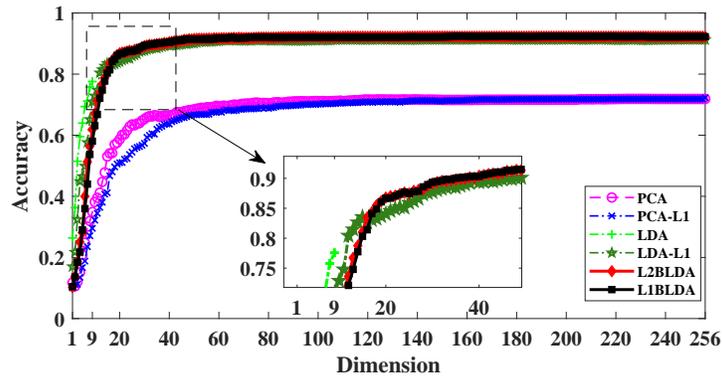}}}\hspace{5pt}
\caption{The variation of accuracies along different dimensions on the MNIST database.}
\label{dimACMNIST}}
\end{center}
\end{figure}

The table shows that for the original data, all the methods behave similarly except for LDA. However, when the samples are contaminated, PCA, PCA-L1 and LDA are all greatly influenced by noise, while LDA-L1 and our L2BLDA, L1BLDA have small changes. In addition, our L2BLDA and L1BLDA are both better than LDA-L1, and our L1BLDA has the best performance. It demonstrates the effectiveness of the proposed methods.
To see how the reduced dimension affect each method, we depict the variation of accuracies along dimensions, as shown in Figure \ref{dimACMNIST}. For the original data, as the dimension grows, the accuracies of all the methods fist grow rapidly and then keep steady with the similar performance. When the noise is considered, our L1BLDA and L2BLDA and LDA-L1 affected less by noise comparing to PCA and PCA-L1, while our L2BLDA and L1BLDA have the higher accuracies than LDA-L1 after dimension 17. This demonstrates the effectiveness of the proposed method.

\subsubsection{The USPS database}
\begin{figure}[htpb]
\begin{center}{
{\resizebox*{9.5cm}{!}
{\includegraphics{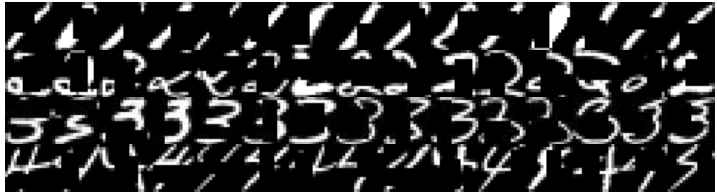}}}\hspace{5pt}
\caption{The contaminated samples from the USPS database.}
\label{USPSsample}}
\end{center}
\end{figure}

\begin{figure}[htpb]
\begin{center}{
\subfigure[Original data]{
\resizebox*{9.5cm}{!}
{\includegraphics{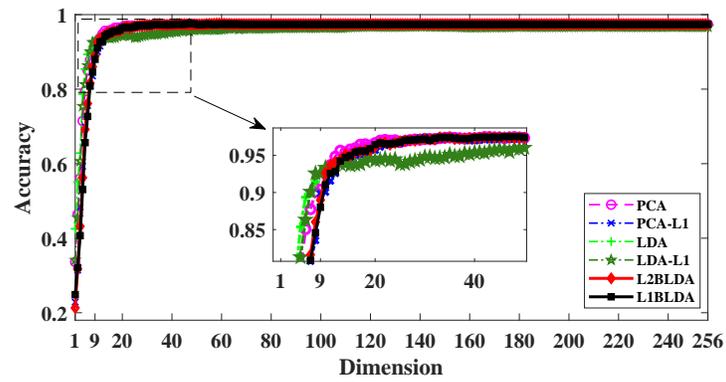}}}\hspace{5pt}
\subfigure[Noise data]{
\resizebox*{9.5cm}{!}
{\includegraphics{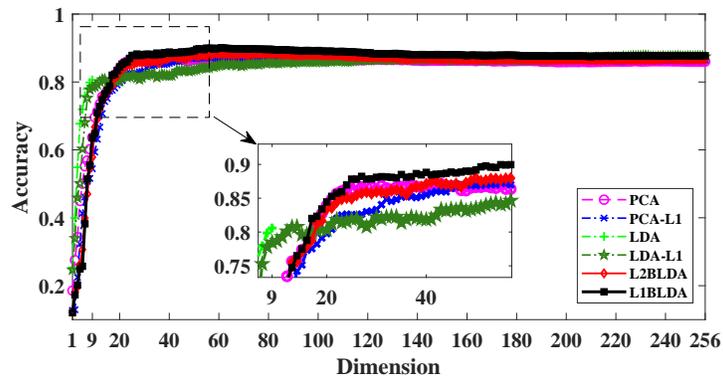}}}\hspace{5pt}
\caption{The variation of accuracies along different dimensions on the USPS database.}
\label{dimACUSPS}}
\end{center}
\end{figure}

The USPS database contains 11000 samples with 10 classes of dimension 256, and each sample corresponds to a digit. We randomly select 80\% samples from each class for training, while the rest data is used for testing. To test the robustness of the proposed method, we further add black block on each training data, where the block covers random 20\% rectangular area of each image, as shown in Figure \ref{USPSsample}.
As before, all the methods are then applied on the original training data and the contaminated training data, and the corresponding results are given in Table \ref{TableUSPS} and Figure \ref{dimACUSPS}. When no noise is added, our L2BLDA performs the best, while our L1BLDA, PCA and PCA-L1 are comparable to L2BLDA. However, when the image is contaminated, L1BLDA behaves the best, which shows its robustness. Similar to the MNIST database, the variation of accuracies along different dimensions shown in Figure \ref{dimACUSPS} also demonstrates the superiority of our proposed methods.
\begin{table*}[htbp]
\begin{center}
\caption{Classification results on the USPS database.}
\resizebox{5.5in}{!}
{
\begin{tabular}{lccccccccccc}
\toprule
{\multirow{2}{*}{Data}}& \vline &PCA&PCA-L1& LDA &LDA-L1&L2BLDA&L1BLDA\\
& \vline& Acc  (Dim)& Acc  (Dim)&Acc  (Dim)& Acc  (Dim)& Acc  (Dim)& Acc  (Dim)\\
\midrule
Original data& \vline&97.53 (55) &97.53 (53)&92.63 (9)&96.88 (139)&\textbf{97.58 (58)} &97.53 (46)\\
Noise data& \vline& 86.86 (55)&86.95 (87)&79.64 (9) & 87.68 (232) &88.41 (80)&\textbf{90.09 (61) }\\
\bottomrule
\end{tabular}
}
\label{TableUSPS}
\end{center}
\end{table*}

\section{Conclusion}\label{secCon}

This paper proposed two novel L1-norm and L2-norm based linear discriminant analysis (L1BLDA and L2BLDA) which were upper bounds of the theoretical framework of the Bhattacharyya optimality. Different from the classical LDA, they both maximize the weighted pairwise between-class scatter and minimize the within-class scatter, while their weighting constants are determined by the involved data set. They both can be solved through simple nongreedy algorithms. The constructions of L1BLDA and L2BLDA make them effective,
and the application of L1-norm makes L1BLDA possess robustness. The experimental results also support their superiority. Our Matlab code can be downloaded from http://www.optimal-group.org/Resources/Code/L1BLDA.html.

\section*{Acknowledgment}

This work is supported by the National Natural Science Foundation of
China (No.61703370 and No.61603338), the Natural Science Foundation of Zhejiang Province (No.LQ17F030003 and No.LY18G010018), and the Natural Science Foundation of Hainan Province (No.118QN181).


%
%
%

\end{document}